\documentclass[lettersize,journal]{IEEEtran}

\usepackage{graphicx}
\usepackage{amsmath}
\usepackage{amssymb}
\usepackage[english]{babel}
\usepackage[utf8]{inputenc}
\usepackage{algorithm}
\usepackage[noend]{algpseudocode}
\usepackage{cleveref}
\usepackage{times}
\usepackage{mathtools}
\usepackage{booktabs}
\usepackage{bm}
\usepackage[dvipsnames]{xcolor}
\usepackage{tikz}
\usepackage[font={footnotesize}]{caption}
\usepackage{gensymb}
\usepackage{subcaption}
\usetikzlibrary{arrows,shapes,automata,backgrounds,petri}
\usetikzlibrary[pgfplots.groupplots]
\usepackage[EULERGREEK]{sansmath}

\usepackage{pgfplots}
\pgfplotsset{compat=newest}
\usetikzlibrary{plotmarks}
\usetikzlibrary{arrows.meta}
\usepgfplotslibrary{patchplots}
\usepackage{grffile}

\newcommand{\bh}{\bm{h}}

\newcommand{\bx}{\bm{x}}
\newcommand{\by}{\bm{y}}

\newcommand{\cN}{N}

\newcommand{\cX}{\mathcal{X}}
\newcommand{\cY}{\mathcal{Y}}

\newcommand{\EE}{\mathbb{E}}

\newcommand{\RR}{\mathbb{R}}

\newcommand{\blambda}{\bm{\lambda}}

\newcommand{\maximize}{\mathop{\mathrm{max}}}
\newcommand{\minimize}{\mathop{\mathrm{min}}}
\newcommand{\maximizewrt}[1]{\mathop{\underset{#1}{\maximize}}}
\newcommand{\minimizewrt}[1]{\mathop{\underset{#1}{\minimize}}}

\newcommand{\norm}[1]{\left\| #1\right\|}

\newcommand{\defeq}{\vcentcolon=}

\newcommand{\Expect}[2]{\EE_{#1}\left[#2\right]}

\newcommand{\codeurl}{https://github.com/intelligent-control-lab/CCHP\_ICRA23}

\colorlet{revision_color}{black}
\newcommand{\revcolor}[1]{\textcolor{revision_color}{#1}}

\begin{document}

\title{Robust and Context-Aware Real-Time Collaborative Robot Handling via Dynamic Gesture Commands}


\author{Rui Chen$^{1}$, Alvin Shek$^{1}$, Changliu Liu$^{1}$
\thanks{$^{1}$Carnegie Mellon University, Pittsburgh, PA. Emails: {\tt\small \{ruic, ashek, cliu6\}@andrew.cmu.edu}}%
}



\maketitle

\begin{abstract}

This paper studies real-time collaborative robot (cobot) handling, where the cobot maneuvers an object under human dynamic gesture commands.
Enabling dynamic gesture commands is useful when the human needs to avoid direct contact with the robot or the object handled by the robot.
However, the key challenge lies in the heterogeneity in human behaviors and the stochasticity in the perception of dynamic gestures, which requires the robot handling policy to be adaptable and robust.
To address these challenges, we introduce \textit{Conditional Collaborative Handling Process} (CCHP) to encode a context-aware cobot handling policy and a procedure to learn such policy from human-human collaboration.
We thoroughly evaluate the adaptability and robustness of CCHP and apply our approach to a real-time cobot assembly task with Kinova Gen3 robot arm.
Results show that our method leads to significantly less human effort and smoother human-robot collaboration than state-of-the-art rule-based approach even with first-time users.

\end{abstract}

\vspace{-12pt}


\section{Introduction}

\begin{figure}
    \centering
    \includegraphics[width=\linewidth]{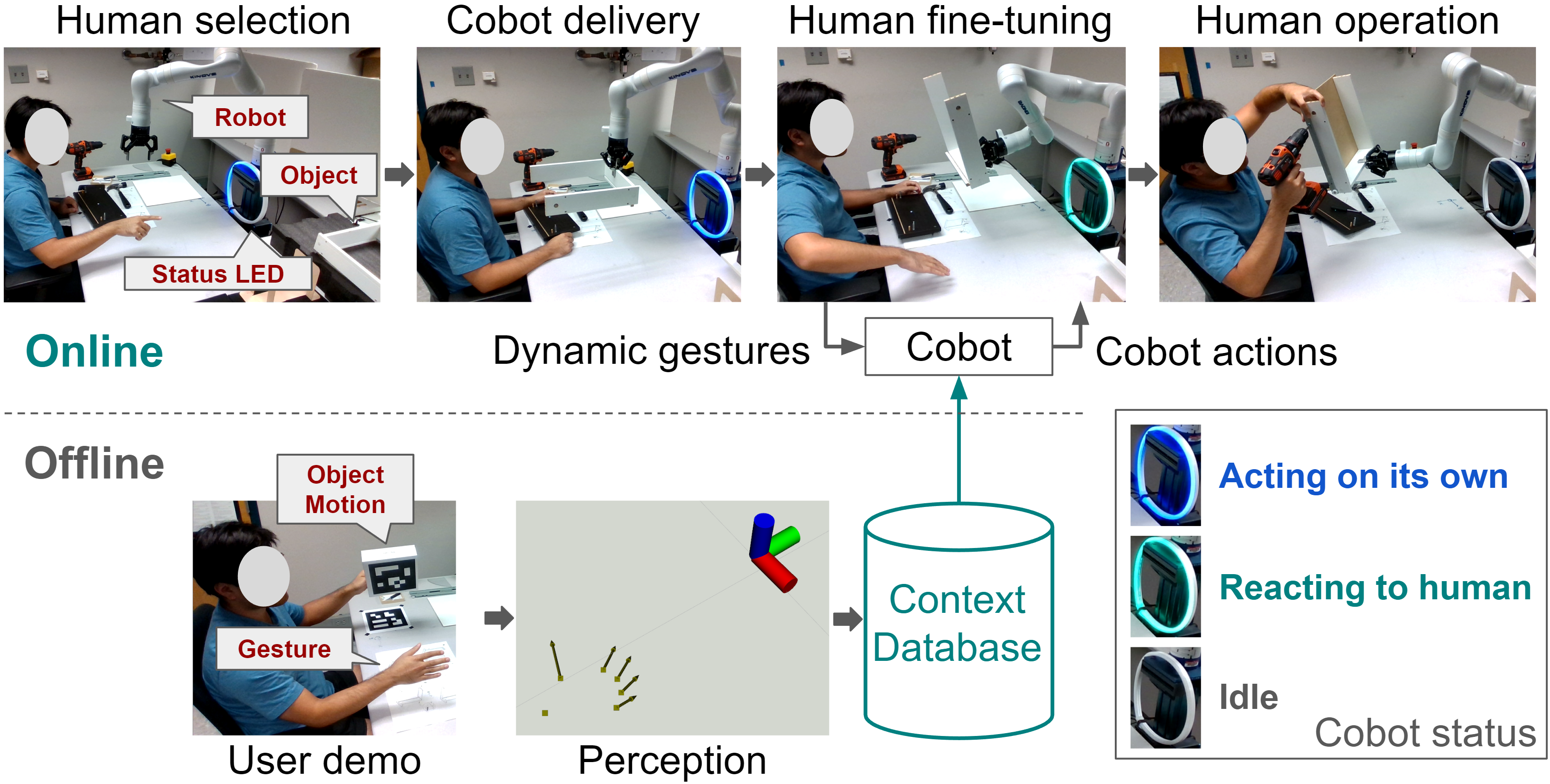}
    \caption{A human collaborates with a robot to assemble a drawer.
    The human adjusts the robot pose for comfortable working condition using dynamic gestures.
    The human commanding style is customized and hinted to the robot via offline demonstrations.
    }
    \label{fig:task_diagram}
    \vspace{-15pt}
\end{figure}

With the advancement of robotic technologies, robots are getting out of cages and directly working with humans.
One typical type of human-robot collaboration (HRC) is \textit{robot-as-tool}~\cite{villani_survey_2018}, meaning that the human handles the cognitive portion (e.g., decision making) and difficult operations (e.g., fit screws inside holes) while the collaborative robot (cobot) provides assistive operations such as tool management (e.g., fetching and returning tools) and object handling (e.g., lifting heavy workpieces for human to work on).
Importantly, although treated as tools, the cobots still run on intelligent algorithms instead of fixed rule-based manners.
A practical challenge is how to make the cobot assistance meet human expectation.
For example, during a surgery, the cobot should always pass a tool with the handle towards the doctor; when helping with furniture assembly, the cobot should always hold furniture pieces at comfortable poses for humans to work on.
However, due to the variation in tasks and human preferences, it is difficult to determine the ``perfect'' cobot handover pose in all situations.
Our insight is that \textit{cobots should allow humans to easily correct their behavior to meet humans' expectations}.
In other words, we aim at the ``last millimeter'' personalized adjustment of cobot pose in robot-as-tool HRC.
We refer to such task as \textit{\textbf{cobot handling}} (see \cref{fig:task_diagram} for an example).
Importantly, to achieve flexible and fluent movements, we allow the cobot end-effector to perform arbitrary rigid body motions from a \textit{continuous action space} as opposed to a discrete set of pre-defined motions.



For cobot handling, hardware interfaces (e.g., joysticks and wearable sensors) are too rigid to perform last millimeter adjustments in a flexible and efficient manner.
Hence, we resort to \textit{natural user interfaces} (NUIs) such as gestures and voices \cite{villani_survey_2018}.
Among NUIs, static gesture is one of the most studied \cite{ende_human_centered_2011,shukla2016,mazhar_real_time_2019,gesture_hri_review,marasovic_motion_based_2015}.
However, in cobot handling where the desired cobot motion changes continuously, static gesture would show limited flexibility due to its discrete nature.
Hence, we propose to use \textbf{dynamic gesture}, since it enables users to directly mimic or depict a mental image of desired object movements, leading to both flexibility and fluidity of interaction.
Then, an immediate question is: how to make cobots understand the dynamic gestures from different human users, and react with handling operations that meet humans' expectation? 

It is challenging in designing robot policies for HRC that can
adapt to different user styles \cite{attitude_2017}, handle uncertainty of natural human input, and generate smooth robot motions.
For example, users might have different joint flexibilities which would render certain gestures natural for some users but hard for others.
It is hence ideal for users to develop their own gesture commands.
To avoid additional training for new users, we should leverage offline user demonstrations (i.e., \textit{context}) alongside online user gestures during test time.
Besides, natural human inputs such as dynamic gestures can be noisy due to human uncertainty and limited quality of the sensing systems.
To capture such uncertainty, we should resort to stochastic solution and learn a distribution of cobot policies.

Integrating the ideas mentioned above, we frame the problem of cobot handling as learning a \textit{distribution of functions conditioned on observations}, where each realization is a policy variant that maps \textit{human dynamic gestures to continuous cobot end-effector motions in real-time}.
To learn the distribution, we propose {\em Conditional Collaborative Handling Process} (CCHP) inspired by a recent line of research on neural processes (NP) \cite{cnp, np, anp}, which has been verified to be adaptable to various data distributions and robust to input uncertainty.
CCHP is implemented and verified on a real-time cobot handling task: collaborative furniture assembly with a Kinova Gen3 robot arm.
User studies show that CCHP leads to less human efforts and better human-robot collaboration with context.

In summary, this paper contributes to the learning and construction of a highly \textit{adaptive} and \textit{robust} robot policy which generates handling operations from a \textit{continuous action space} in \textit{real-time} with human commands.
The rest of this paper is organized as follows.
In \cref{sec:related}, we review literature related to gesture-based human-robot interaction as well as functional learning. In \cref{sec:problem}, we formulate the problem of cobot handling.
In \cref{sec:cchp}, we derive CCHP and describe its implementation.
In \cref{sec:eval}, we present qualitative and quantitative evaluation of CCHP.
In \cref{sec:user_study}, we present user studies on a realistic HRC task.
Finally, we summarize this work and discuss about future directions in \cref{sec:conclusion}.

\vspace{-10pt}

\section{Related Work}
\label{sec:related}



\subsection{Gesture-Based Human Robot Interaction}

In literature, there are two major forms of gesture-based commands: \textit{static gestures} and \textit{dynamic gestures}.
Regarding static gestures, the most common approach is to classify gesture poses using a finite set of symbolic labels which are further mapped to robot actions \cite{ende_human_centered_2011, mazhar_real_time_2019}.
Such an approach is unsuitable for cobot handling tasks that are inherently continuous.
Regarding dynamic gestures, \cite{marasovic_motion_based_2015} treats the whole hand as a single point and interprets its trajectory using some simple geometries such as circles and alphabetical letters.
Other approaches recognize dynamic gestures using feature-based template matching \cite{depth_sensor_material_handle} or convolutional neural networks \cite{rt_multi_model_HRC_gesture_speech}.
However, those approaches still operate on a discrete set of pre-defined robot actions.
To fully leverage the expressiveness of dynamic gestures and achieve seamless human-robot interaction, the robot should react to human gestures in real time with continuous actions.
Such setting introduces significant challenge in cobot policy because of the continuous robot action space, and existing methods fail to solve the problem in flexible and adaptable fashion.
For example, \cite{hand_to_end_effector} maps hand motions to real-time robot gripper actions based on hand keypoint detection.
This approach enforces a pre-designed hand gesture which may appear hard to users with limited strength and joint flexibilities.
Another work \cite{fan_object_2016} achieves real-time object pose tracking from hand motions.
Their approach requires costly motion capture systems and wearable markers and hence is not flexible.
Also, they do not consider different hand motion styles from different users.
For further review, we refer readers to \cite{FAN2022102304} for a wide variety of other recognition strategies.
In this paper, we desire a cobot handler that can be easily commanded by various users in real time without wearing any device.
This requires the cobot to adapt to different user control strategies and hand gesture patterns, which is not solved by existing approaches.

\vspace{-12pt}



\subsection{Functional Distribution with Efficient Inference}

The goal of cobot handling is to learn a policy distribution that models human uncertainty, while being able to adapt to different users based on user-specific demonstrations.
On an abstract level, this is equivalent to regressing a functional distribution which predicts function values at unobserved input locations with uncertainty, given some previous observations.
One direct approach is to perform inference on a stochastic process such as Gaussian process (GP).
However, GP's computation complexity \cite{gp_quad} is too high for real-time tasks which require fast online inferences.
Moreover, it is hard to design the GP kernel functions for high-dimensional tasks like ours.

There is a line of research that models stochastic processes with a class of neural networks, named \textit{neural processes} (NP), to achieve linear computational complexity with respect to observations during test time.
This approach is first formally presented as conditional neural processes (CNP) \cite{cnp} which explicitly incorporate training data at test time as observations.
Subsequent improvements \cite{np, anp} have been effective on tasks such as image completion.
However, they cannot be directly applied to cobot handling which is significantly more complex and higher dimensional.
More importantly, our task output (handling operations) should carry strong temporal structures, which are not considered by NPs.

To define stochastic processes, we indeed need to ensure invariance to input permutations, i.e., exchangeability condition \cite{Oksendal2003}.
However, it is reported to be practically beneficial to relax such assumption when the observations contain time sequences \cite{qin_recurrent_2019}.
Specifically, recurrent attentive neural processes (RANP) \cite{qin_recurrent_2019} incorporates a recurrent neural network structure to process the observations, and show improved performance on vehicle trajectory predicition over ANP.
In RANP, the exchangeability is only relaxed on observations, while the temporal structure of test-time input and output is not considered.
In cobot handling, we also need to relax exchangeability condition at test time to ensure smooth and consistent cobot actions.
Finally, there are other extensions to NPs \cite{rnp,snp,robust_snp} whose problem setting deviates from ours.





\vspace{-10pt}
\section{Mathematical Problem for Cobot Handling}
\label{sec:problem}


\subsection{Real-time cobot handling task and terminology}
\label{sec:rtcohand_terminology}

In cobot handling, we assume that an object of interest is held by the cobot end-effector, and the user adjusts the object pose via dynamic gesture commands.
In this paper, we focus on only the right hand for simplicity, while our framework can be extended to using both hands directly.
We first define dynamic gestures as sequences of full hand poses, denoted by $\bx\defeq \left(x^{(1)}, x^{(2)}, \dots, x^{(N)}\right)\in\cX^N$.
Each $x^{(t)}\in \cX$ encodes the \revcolor{3D locations of $21$} hand skeleton keypoints at time $t$ and $\cX\subset\RR^{D_x}$ ($D_x=63$).
$\cX^N \defeq \cX\times\cX\dots\times\cX$ is the dynamic gesture space.
Since it is more natural for users to command relative robot motions rather than absolute poses, we define the cobot handling operation as a sequence of $6\mathrm{D}$ rigid body velocities $y\in\cY$ of the end-effector defined in Cartesian space.
$\cY\subset\RR^{D_y}$ is the space of all possible Cartesian velocities (i.e., $D_y=6$).
We then denote handling operation (e.g., object motion) as $\by\defeq \left(y^{(1)}, y^{(2)}, \dots, y^{(N)}\right)\in\cY^N$ where $\cY^N \defeq \cY \times \cY\dots\times\cY$ is the operation space.
Here, $N$ refers to the duration of any continuous cobot handling session and can vary as needed.
To command each desired operation $\by$ on the object, users perform a dynamic gesture $\bx$.
To ease analysis, we assume both sequences to have the same length.
In practice, the generated handling operation can be interpolated or sub-sampled if denser or sparser control is desired.

We formulate our problem as constructing a cobot policy $\pi_\theta:\cX^N\mapsto\cY^N$ that maps human command $\bx$ to handling operation $\by$, where $\theta$ is the parameter.
This policy should achieve human-level assistance as if a human helper were performing operations that match the users' expectations.
This can be written as the following minimization
\begin{equation}\label{eq:cobot_goal}
\minimizewrt{\theta} \Expect{(\bx,\by)\sim D_\mathsf{train}}{\norm{\pi_\theta(\bx)-\by}}.
\end{equation}
where $D_\mathsf{train}$ is a human-labeled dataset and $\norm{\cdot}$ is the norm function.
For fluent collaboration, we add a real-time requirement that the cobot computes and applies handling operation $\by$ at the same time when users perform commands $\bx$.
See \cref{fig:task_diagram} for an illustration of cobot handling task.
\vspace{-10pt}

\begin{figure}[t]
    \centering
    \includegraphics[width=0.8\linewidth]{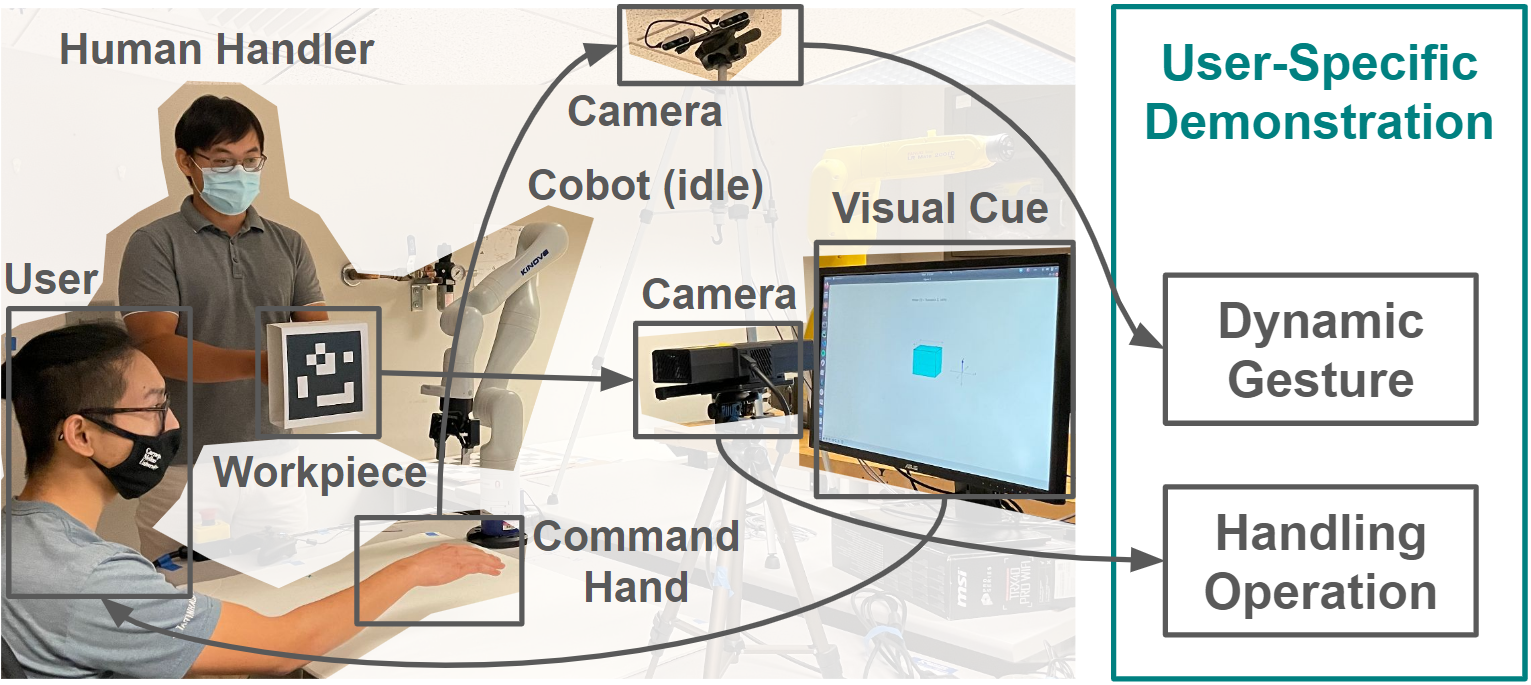}
    \caption{Creating demonstrations for cobot handling from human-human collaboration.
    The user performs dynamic gestures to achieve the object motions as indicated by the visual cue.
    The handling operations are labeled by a \revcolor{human handler} using a trackable object.
    The dynamic gestures are detected using OpenPose\cite{openpose} library with dual Intel RealSense cameras and saved as 3D locations of hand skeleton key points.
    The object poses are recorded by tracking an ArUco marker attached to it and saved as 6D Cartesian poses.}
    \label{fig:rtcohand_info}
    \vspace{-15pt}
\end{figure}

\subsection{Addressing user adaptation and human uncertainty}\label{sec:formulation_derive}

Due to the difference in strength and flexibility, the same dynamic gesture can appear natural and intuitive for certain users but not for others.
For example, rotating the wrist without moving the arm might only appear easy to users with high wrist flexibility.
Hence, we introduce \textit{policy customization} and let users design their own dynamic gestures.
When collecting $D_\mathsf{train}$, each user demonstrates the handling task with another \revcolor{\textit{human handler}} acting in the role of cobot in real-time (see \cref{fig:rtcohand_info}).
The \revcolor{human handler} can communicate with the user to ensure correct understanding of the dynamic gestures.

\paragraph*{\textbf{Challenge of adapting to different or new users}}

\begin{figure}[t]
    \centering
    \begin{subfigure}[b]{0.42\linewidth}
        \centering
        \resizebox{1.0\linewidth}{!}{\begin{tikzpicture}[node distance=1.2cm, thin, -latex, bend angle=45, auto]

    \definecolor{ContextColor}{RGB}{56,142,142}
    \colorlet{TrainColor}{Black}

    \tikzstyle{obs} = [circle, text centered, minimum size=1.0cm, draw=black, fill=black!10, semithick]
    \tikzstyle{latent} = [circle, text centered, minimum size=1.0cm, draw=black, fill=black!0, semithick]
    \tikzstyle{label} = [rectangle, minimum width=0.1cm, minimum height=0.1cm, text centered]

    \tikzstyle{entity} = [rectangle, rounded corners, minimum width=0.1cm, minimum height=0.6cm, text centered]
    \tikzstyle{entity_mid} = [rectangle, rounded corners, minimum width=0.1cm, minimum height=0.6cm, text centered]

    \tikzstyle{arrow} = [semithick,->,-latex]
    \tikzstyle{arrow_train} = [semithick,->,-latex,draw=TrainColor]
    \tikzstyle{arrow_test} = [semithick,->,-latex,draw=ContextColor]

    \node [entity] (user) {$\mathsf{User}$};

    \node [latent] (x_train) [at=(user.west |- user.east), xshift=-1.0cm] {$\bx_{\mathsf{train}}$};
    \node [entity, align=center, font=\footnotesize] (human_expert) [below of=x_train] {$\mathsf{Human}$\\$\mathsf{Handler}$};
    \node [latent] (y_train) [below of=human_expert] {$\by_{\mathsf{train}}$};

    \draw [arrow] (user) to (x_train);
    \draw [arrow] (x_train) to (human_expert);
    \draw [arrow] (human_expert) to (y_train);

    \node [entity_mid, font=\footnotesize, text=TrainColor] (learn) [at=(human_expert -| user.south), yshift=0.5cm] {$\mathsf{Learn}$};
    \node [entity_mid, font=\footnotesize, text=ContextColor] (dep) [at=(human_expert -| user.south), yshift=-0.5cm] {$\mathsf{Context}$};

    \node [latent] (x_test) [at=(user.east |- user.east), xshift=1.0cm] {$\bx_{\mathsf{test}}$};
    \node [entity, align=center, font=\footnotesize] (cobot_policy) [below of=x_test] {$\mathsf{Cobot}$\\$\mathsf{Policy}$};
    \node [latent] (y_test) [below of=cobot_policy] {$\by_{\mathsf{test}}$};

    \draw [arrow] (user) to (x_test);
    \draw [arrow] (x_test) to (cobot_policy);
    \draw [arrow] (cobot_policy) to (y_test);

    \draw [arrow_train] (x_train) to (learn.west);
    \draw [arrow_train] (y_train) to (learn.west);
    \draw [arrow_train] (learn.east) to (cobot_policy.west);

    \draw [arrow_test] (x_train) to (dep.west);
    \draw [arrow_test] (y_train) to (dep.west);
    \draw [arrow_test] (dep.east) to (y_test.west);


    \begin{pgfonlayer}{background}
        \filldraw [line width=3mm,join=round,black!20]
            (x_train.north  -| x_train.east) rectangle (y_train.south  -| y_train.west)
            (x_test.north  -| x_test.east) rectangle (y_test.south  -| y_test.west);
    \end{pgfonlayer}

    \node [label, font=\footnotesize, align=center] (pc) [above of=x_train, yshift=-0.1cm] {$\mathsf{Policy~Customization}$\\$\mathsf{(training)}$};
    \node [label, font=\footnotesize, align=center] (deployment) [above of=x_test, yshift=-0.1cm] {$\mathsf{Deployment}$\\$\mathsf{(testing)}$};

\end{tikzpicture}}
        \caption{\footnotesize Train test phases.}
    \label{fig:train_test_procedure}
    \end{subfigure}
    \hfill
    \begin{subfigure}[b]{0.56\linewidth}
        \centering
        \resizebox{1.0\linewidth}{!}{\begin{tikzpicture}[node distance=1.4cm, thin, -latex, bend angle=45, auto]

    \definecolor{ContextColor}{RGB}{56,142,142}
    \definecolor{TargetColor}{HTML}{CC4125}

    \tikzstyle{obs} = [circle, text centered, minimum size=1.0cm, draw=black, fill=black!10, semithick]
    \tikzstyle{latent} = [circle, text centered, minimum size=1.0cm, draw=black, fill=black!0, semithick]
    \tikzstyle{label} = [rectangle, minimum width=0.1cm, minimum height=0.1cm, text centered]

    \tikzstyle{arrow} = [semithick,->,-latex]
    \tikzstyle{arrow_bg} = [semithick,->,-latex,draw=black!100]
    
    \node [obs] (yT1) {$y_T$};
    \node [obs] (yT2) [right of=yT1] {$y_T$};
    \node [label] (dots_y) [right of=yT2] {$\dots$};
    \node [obs] (yTN) [right of=dots_y] {$y_T$};
    \draw [arrow] (yT1) to (yT2);
    \draw [arrow] (yT2) to (dots_y);
    \draw [arrow] (dots_y) to (yTN);

    \node [latent] (z) [below of=yT2, xshift=0.8cm, yshift=0.3cm] {$z$};

    \node [obs] (xT1) [above of=yT1, yshift=0.5cm] {$x_T$};
    \node [obs] (xT2) [above of=yT2, yshift=0.5cm] {$x_T$};
    \node [label] (dots_x) [above of=dots_y, yshift=0.5cm] {$\dots$};
    \node [obs] (xTN) [above of=yTN, yshift=0.5cm] {$x_T$};

    \draw [arrow] (xT1) to (yT1);
    \draw [arrow] (xT2) to (yT2);
    \draw [arrow] (xTN) to (yTN);

    \draw [arrow] (z) to (yT1.south);
    \draw [arrow] (z) to (yT2.south);
    \draw [arrow] (z) to (yT2.south -| dots_y.south);
    \draw [arrow] (z) to (yTN.south);

    \node [label] (t1) [above of=xT1, yshift=-0.6cm] {$t=1$};
    \node [label] (t2) [above of=xT2, yshift=-0.6cm] {$t=2$};
    \node [label] (tN) [above of=xTN, yshift=-0.6cm] {$t=N_T$};

    \node [obs] (xC) [left of=xT1, xshift=-0.2cm, yshift=-0.5cm] {$x_C$};
    \node [obs] (yC) [left of=yT1, xshift=-0.2cm] {$y_C$};

    \draw [join=round,black,rounded corners,semithick]
            ([xshift=-0.2cm, yshift=0.2cm] xC.west |- xC.north) rectangle ([xshift=0.2cm, yshift=-0.4cm] yC.south  -| yC.east);
    \node [label] (NC) [below of=yC, xshift=0.4cm, yshift=0.7cm] {$N_C$};

    \draw [arrow_bg] (xC) to[out=-5,in=135] (yT1.north west);
    \draw [arrow_bg] (yC) to[out=35,in=135] (yT1.north west);

    \draw [arrow_bg] (xC) to[out=-5,in=140] (yT2.north west);
    \draw [arrow_bg] (yC) to[out=35,in=140] (yT2.north west);

    \draw [arrow_bg] (xC) to[out=-3,in=160] (yTN.north west);
    \draw [arrow_bg] (yC) to[out=35,in=160] (yTN.north west);

    \node [label] (context_label) [above of=xC, yshift=-0.4cm, text=ContextColor] {$\mathsf{Context}$};
    \node [label] (target_label) [above of=xT2, yshift=-0.2cm, xshift=0.8cm, text=TargetColor] {$\mathsf{Target}$};


\end{tikzpicture}}
        \caption{\footnotesize Graphical model of CCHP.}
    \label{fig:pgm}
    \end{subfigure}
    \caption{\footnotesize (a) computation diagram of CCHP.
    (b) graphical model of CCHP.
    }
    \label{fig:procedure_pgm}
    \vspace{-15pt}
\end{figure}
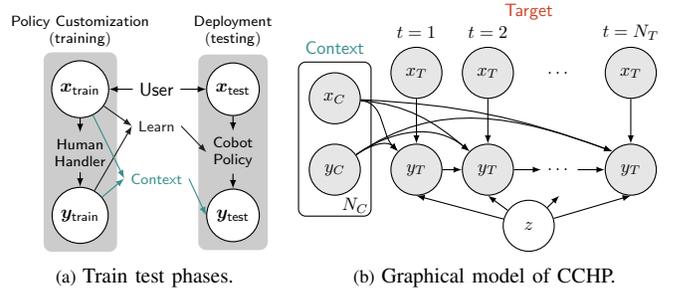

With policy customization, the dynamic gestures $\bx$ for each handling operation $\by$ from different users can vary.
Hence, we need to adapt efficiently without altering the robot policy (e.g., via fine-tuning) for new users.
To achieve this, we assume access to some prior knowledge about users and introduce explicit dependence of $\pi_\theta$ on it during runtime.
The prior is essentially some annotated command-operation pairs $\{(\bx_C,\by_C)\}$, referred to as $\textit{context}$.
Then, we refer to new user commands and true desired operations as $\textit{target}$, denote as $\{(\bx_T,\by_T)\}$.
In this spirit, we update the goal of cobot handling \eqref{eq:cobot_goal} with a conditional form as
\begin{equation}\label{optim:cobot_goal_2}
    \minimizewrt{\theta} \Expect{(\bx_T,\by_T)\sim D_{\mathsf{test}}}{\norm{\pi_\theta(\bx_T\mid \bx_C,\by_C) - \by_T}}.
\end{equation}
All $(\bx,\by)$ pairs are associated with the human demonstrator ID, so that we can match the context with user during testing.
In this way, $\pi_\theta$ no longer needs to fully encode each user's preferences, since it can draw insights from user-specific database.
Now, notice that \eqref{optim:cobot_goal_2} represents a testing (deployment) time objective and cannot be directly solved, because the true desired operation $\by_T$ is unavailable during training.
Hence, at training time, we split the demonstration $D_{\mathsf{train}}$ into context and target to simulate the testing scenario, yielding the following formulation
\begin{equation}\label{optim:cobot_goal_2_train}
    \minimizewrt{\theta} \Expect{(\bx_C,\by_C,\bx_T,\by_T)\sim D_{\mathsf{train}}}{\norm{\pi_\theta(\bx_T\mid \bx_C,\by_C) - \by_T}}.
\end{equation}
See \cref{fig:train_test_procedure} for an illustration of the training and testing phases.
Our idea of solving adaptation via conditional prediction is partially inspired by the line of research on NPs \cite{cnp,np,anp} and conceptually similar to few-shot learning, where the target data is compared to observed data in some feature space \cite{siamese, vinyals_matching_2017, bartunov_fast_2017}. For more detailed analysis of such connection, we refer readers to \cite{cnp,np,anp}.

\paragraph*{\textbf{Challenge of modeling human motion uncertainties}}

During training and testing,
the actual dynamic gestures $\bx$ carried out by users for the same desired operation $\by$ might vary among multiple trials, \revcolor{and vice versa}.
To capture such uncertainty, we adopt a stochastic version of the handling policy $\Pi_\theta$ such that $\Pi_\theta(\bx_T)\sim p_\theta(\by_T\mid\bx_T,\bx_C,\by_C)$.
$p_\theta$ is a conditional probability modeling the human-human demonstration.
With that in hand, we arrive at our final formulation of the cobot handling goal as learning a conditional distribution:
\begin{equation}\label{optim:cobot_goal_3}
    \maximizewrt{\theta} \Expect{(\bx_C,\by_C,\bx_T,\by_T)\sim D_\mathsf{train}}{\log p_\theta(\by_T\mid\bx_T,\bx_C,\by_C)}.
\end{equation}




Notably, we acquire the best model when the user command style is consistent between context and during testing.
Namely, the $\bx$'s should be the same or at least similar for the same $\by$'s and distinguished for different $\by$'s.
Otherwise, the model can hardly learn to relate the target to the context and will lose the desired adaptation power.
In this work, we reject user demonstrations with inconsistent styles and leave the problem of non-stationary styles for future work.
The measurement of styles and rejection criteria will be described in \cref{sec:collect_user_policies}.
Next, we further model $p_\theta$ combining our insights on cobot handling and practically solve \eqref{optim:cobot_goal_3}.

\vspace{-10pt}

\section{Conditional Collaborative Handling Process}
\label{sec:cchp}



In this section, we formally propose {\em conditional collaborative handling process} (CCHP) to model the cobot handling policy $\Pi_\theta$.
Then, we derive the learning objective of CCHP for solving our goal \eqref{optim:cobot_goal_3} with a brief description of the implementation.
Note that our approach also applies to other general robotic tasks where offline context is available, with minor changes for accommodating the actual task input and output.

\vspace{-15pt}

\subsection{Probabilistic Perspective of Cobot Handling}\label{sec:cchp_intro}

We desire a stochastic cobot handling policy $\Pi_\theta$ which samples from a probability distribution conditioning on user-specific prior, i.e., $\Pi_\theta(\bx_T)\sim p_\theta(\by_T\mid\bx_T,\bx_C,\by_C)$.
To model this distribution, we propose {\em conditional collaborative handling process} (CCHP).
See \cref{fig:pgm} for a graphical representation.
The CCHP incorporates two key features, each attending to a practical challenge (see section \ref{sec:formulation_derive}): (a) the explicit dependence of $\Pi_\theta$ on a user-specific context $(\bx_C,\by_C)$ for user adaptation and (b) a latent variable $z\in\RR^{D_z}$ that captures the underlying randomness of handling operations $\by_T$.
Intuitively, $z$ can encode a wide range of characteristics of handling operations, e.g., how dynamic gesture patterns map to those in handling operations, the overall scale of cobot movements, and the amount of uncertainty.
\revcolor{
A larger $D_z$ is needed for more complex robot policies, and $D_z=32$ in this work achieves promising results.
}
Importantly, we also introduce temporal dependency between timesteps in $\by_T$ since the handling operations should not change too fast even if the input is noisy.
The generative process can be written as
\begin{align}\label{eq:generative_process}
    & p_\theta(\by_T\mid \bx_T,\bx_C,\by_C) \defeq \int p_\theta(\by_T\mid \bx_T,\bx_C,\by_C,z) p(z) dz. \nonumber \\
    & = \int \prod_{t=1}^{N_T} p_\theta(y_T^{(t)} \mid \by_T^{(1:t-1)}, x_T^{(t)},\bx_C,\by_C,z) p(z) dz.
\end{align}
where $\by_T^{(1:t-1)}\defeq (y_T^{(1)},y_T^{(2)},\dots,y_T^{(t-1)})$.

\vspace{-10pt}



\begin{figure*}[ht]
    \centering

    \begin{subfigure}[b]{0.8\linewidth}
        \centering
        \input{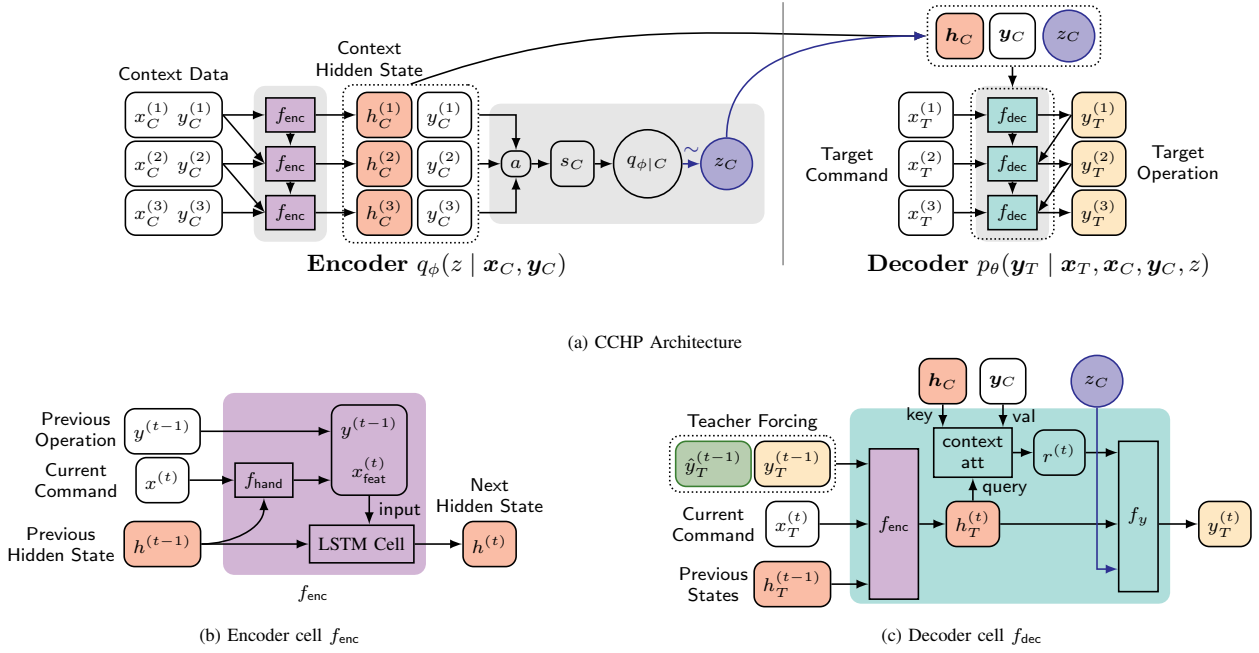}
        \caption{CCHP Architecture}
        \label{fig:CCHP_arch}
    \end{subfigure}\\
    \begin{subfigure}[b]{0.45\linewidth}
        \centering
        \colorlet{TrainColor}{OliveGreen}
\colorlet{TestColor}{Blue}
\colorlet{TrainColorFill}{TrainColor!30}
\colorlet{TestColorFill}{TestColor!30}

\colorlet{PredColor}{Dandelion}
\colorlet{PredColorFill}{PredColor!30}

\colorlet{HiddenColor}{Red!30}
\colorlet{zPriorColor}{TestColorFill}
\colorlet{zPostColor}{TrainColorFill}
\colorlet{FACColor}{Plum!30}
\colorlet{DecCellColor}{Emerald!30}

\begin{tikzpicture}[node distance=0.65cm, thin, -latex, bend angle=45, auto, font=\scriptsize]

    \tikzstyle{var} = [rectangle, rounded corners, minimum width=0.1cm, minimum height=0.6cm, text centered, draw=black, semithick]

    \tikzstyle{var_hidden} = [rectangle, rounded corners, minimum width=0.1cm, minimum height=0.6cm, text centered, draw=black, fill=HiddenColor, semithick]

    \tikzstyle{var_train} = [rectangle, rounded corners, minimum width=0.1cm, minimum height=0.6cm, text centered, draw=TrainColor, semithick, fill=TrainColorFill]

    \tikzstyle{var_pred} = [rectangle, rounded corners, minimum width=0.1cm, minimum height=0.6cm, text centered, draw=black, semithick, fill=PredColorFill]

    \tikzstyle{select} = [rectangle, rounded corners, text centered, draw=black,semithick, densely dotted]

    \tikzstyle{dist} = [circle, text centered, draw=black,semithick]

    \tikzstyle{realis_prior} = [circle, text centered, draw=TestColor,semithick,fill=zPriorColor]

    \tikzstyle{realis_post} = [circle, text centered, draw=TrainColor,semithick,fill=zPostColor]

    \tikzstyle{feat} = [rectangle, minimum width=0.4cm, minimum height=1.7cm,text centered, draw=black,semithick]

    \tikzstyle{cell} = [rectangle, minimum width=0.1cm, minimum height=0.1cm, text centered, draw=black,semithick]

    \tikzstyle{cell_train} = [rectangle, minimum width=0.1cm, minimum height=0.1cm, text centered, draw=TrainColor,semithick, fill=TrainColorFill]

    \tikzstyle{aggre} = [rectangle, rounded corners, minimum width=0.1cm, minimum height=0.1cm, text centered, draw=black,semithick]

    \tikzstyle{label} = [rectangle, minimum width=0.1cm, minimum height=0.1cm, text centered]

    \tikzstyle{arrow} = [semithick,->,-latex]
    \tikzstyle{arrow_train} = [semithick,->,-latex,draw=TrainColor]
    \tikzstyle{arrow_test} = [semithick,->,-latex,draw=TestColor]

    \begin{scope}

        \node [var] (x_feat_t) {$x^{(t)}$};
        \node [cell] (FA) [right of=x_feat_t, xshift=0.7cm] {$f_\mathsf{hand}$};
        \node [var_hidden] (h_tm1) [below of=x_feat_t, yshift=-0.2cm] {$h^{(t-1)}$};
        \node [var] (y_feat_tm1) [above of=x_feat_t, yshift=0.0cm] {$y^{(t-1)}$};

        \node [label, align=center] (yt_1_label) [left of=y_feat_tm1, xshift=-0.5cm] {$\mathsf{Previous}$\\$\mathsf{Operation}$};
        \node [label, align=center] (xt_1_label) [left of=x_feat_t, xshift=-0.5cm] {$\mathsf{Current}$\\$\mathsf{Command}$};
        \node [label, align=center] (ht_1_label) [left of=h_tm1, xshift=-0.7cm] {$\mathsf{Previous}$\\$\mathsf{Hidden~State}$};

        \draw [arrow] (x_feat_t.east) -- (FA.west);
        \draw [arrow] (h_tm1.east) to [out=0,in=-90] (FA.south);
        \node [var, align=center, minimum height=1.2cm] (x_feat_att_t) [right of=FA, xshift=0.75cm, yshift=0.4cm] {$y^{(t-1)}$\\~\\$x^{(t)}_{\mathsf{feat}}$};
        \draw [arrow] (FA.east) -- (x_feat_att_t.west |- FA.east);
        \draw [arrow] (y_feat_tm1.east) -- (x_feat_att_t.west |- y_feat_tm1.east);
        
        \node [cell, minimum height=0.5cm] (lstm_cell) [right of=h_tm1, xshift=2.0cm] {LSTM Cell};
        \draw [arrow] (h_tm1.east) -- (h_tm1.east -| lstm_cell.west);

        \draw [arrow] (x_feat_att_t.south) -- node[right] {$\mathsf{input}$} (lstm_cell.north -| x_feat_att_t.south);

        \node [var_hidden] (h_t) [right of=h_tm1, xshift=3.7cm] {$h^{(t)}$};
        \node [label, align=center] (ht_label) [above of=h_t] {$\mathsf{Next}$\\$\mathsf{Hidden~State}$};
        \draw [arrow] (h_t.west -| lstm_cell.east) -- (h_t.west);

        \node [label] (encoder_cell_label) [below of=lstm_cell, yshift=-0.0cm, xshift=-0.65cm] {$f_\mathsf{enc}$};
        
    \end{scope}

    \begin{pgfonlayer}{background}
        \filldraw [line width=3mm,join=round,FACColor]
            (x_feat_att_t.north  -| lstm_cell.east) rectangle (lstm_cell.south  -| FA.west);
    \end{pgfonlayer}

\end{tikzpicture}
        \caption{Encoder cell $f_\mathsf{enc}$}
        \label{fig:enc_arch}
    \end{subfigure}%
    \begin{subfigure}[b]{0.55\linewidth}
        \centering
        \colorlet{TrainColor}{OliveGreen}
\colorlet{TestColor}{Blue}
\colorlet{TrainColorFill}{TrainColor!30}
\colorlet{TestColorFill}{TestColor!30}

\colorlet{PredColor}{Dandelion}
\colorlet{PredColorFill}{PredColor!30}

\colorlet{HiddenColor}{Red!30}
\colorlet{zPriorColor}{TestColorFill}
\colorlet{zPostColor}{TrainColorFill}
\colorlet{FACColor}{Plum!30}
\colorlet{DecCellColor}{Emerald!30}

\begin{tikzpicture}[node distance=0.65cm, thin, -latex, bend angle=45, auto, font=\scriptsize]

    \tikzstyle{var} = [rectangle, rounded corners, minimum width=0.1cm, minimum height=0.6cm, text centered, draw=black, semithick]

    \tikzstyle{var_hidden} = [rectangle, rounded corners, minimum width=0.1cm, minimum height=0.6cm, text centered, draw=black, fill=HiddenColor, semithick]

    \tikzstyle{var_train} = [rectangle, rounded corners, minimum width=0.1cm, minimum height=0.6cm, text centered, draw=TrainColor, semithick, fill=TrainColorFill]

    \tikzstyle{var_pred} = [rectangle, rounded corners, minimum width=0.1cm, minimum height=0.6cm, text centered, draw=black, semithick, fill=PredColorFill]

    \tikzstyle{select} = [rectangle, rounded corners, text centered, draw=black,semithick, densely dotted]

    \tikzstyle{dist} = [circle, text centered, draw=black,semithick]

    \tikzstyle{realis_prior} = [circle, text centered, draw=TestColor,semithick,fill=zPriorColor]

    \tikzstyle{realis_post} = [circle, text centered, draw=TrainColor,semithick,fill=zPostColor]

    \tikzstyle{feat} = [rectangle, minimum width=0.4cm, minimum height=1.7cm,text centered, draw=black,semithick]

    \tikzstyle{cell} = [rectangle, minimum width=0.1cm, minimum height=0.1cm, text centered, draw=black,semithick]

    \tikzstyle{cell_train} = [rectangle, minimum width=0.1cm, minimum height=0.1cm, text centered, draw=TrainColor,semithick, fill=TrainColorFill]

    \tikzstyle{aggre} = [rectangle, rounded corners, minimum width=0.1cm, minimum height=0.1cm, text centered, draw=black,semithick]

    \tikzstyle{label} = [rectangle, minimum width=0.1cm, minimum height=0.1cm, text centered]

    \tikzstyle{arrow} = [semithick,->,-latex]
    \tikzstyle{arrow_train} = [semithick,->,-latex,draw=TrainColor]
    \tikzstyle{arrow_test} = [semithick,->,-latex,draw=TestColor]

    \begin{scope}

        \node [var_train] (y_feat_T_tm1) {$\hat{y}^{(t-1)}_T$};
        \node [var_pred] (y_feat_pred_tm1) [right of=y_feat_T_tm1, xshift=0.4cm] {$y^{(t-1)}_T$};
        \node [select, minimum height=0.75cm, minimum width=2.2cm, xshift=-0.13cm] (tf_sel) [right of=y_feat_T_tm1] {};
        \node [label] (tf_label) [above of=tf_sel, yshift=-0.1cm] {$\mathsf{Teacher~Forcing}$};
        \node [var] (x_feat_T_t) [below of=y_feat_pred_tm1, xshift=0.0cm, yshift=-0.15cm] {$x^{(t)}_T$};
        \node [var_hidden] (h_tm1_dec) [below of=x_feat_T_t, yshift=-0.15cm] {$h^{(t-1)}_T$};
        
        \node [label] (x_feat_T_t_label) [left of=x_feat_T_t, xshift=-0.4cm, align=center] {$\mathsf{Current}$\\$\mathsf{Command}$};
        \node [label] (h_t_1_label) [left of=h_tm1_dec, xshift=-0.4cm, align=center] {$\mathsf{Previous}$\\$\mathsf{States}$};

        \node [feat, minimum height=2.0cm, fill=FACColor] (dec_fac) [right of=x_feat_T_t, xshift=0.7cm, yshift=0.0cm] {$f_\mathsf{enc}$};
        \draw [arrow] (tf_sel.east) -- (tf_sel.east -| dec_fac.west);
        \draw [arrow] (h_tm1_dec.east) -- (h_tm1_dec.east -| dec_fac.west);
        \draw [arrow] (x_feat_T_t.east) -- (x_feat_T_t.east -| dec_fac.west);

        \node [var_hidden] (h_t_dec) [right of=dec_fac, xshift=0.4cm] {$h^{(t)}_T$};
        \draw [arrow] (h_t_dec.west -| dec_fac.east) -- (h_t_dec.west);

        \node [cell, align=center] (context_att) [above of=h_t_dec, xshift=0.0cm, yshift=0.3cm] {$\mathsf{context}$\\$\mathsf{att}$};
        \node [var_hidden] (hC) [above of=context_att, xshift=-0.4cm, yshift=0.3cm] {$\bh_C$};
        \node [var] (y_feat_C) [above of=context_att, xshift=0.4cm, yshift=0.3cm] {${\by}_C$};
        \draw [arrow] (hC.south) -- node[left]{$\mathsf{key}$} (hC.south |- context_att.north);
        \draw [arrow] (y_feat_C.south) -- node[right]{$\mathsf{val}$} (y_feat_C.south |- context_att.north);
        \draw [arrow] (h_t_dec.north) to node[right, xshift=0.0cm, yshift=0.0cm]{$\mathsf{query}$} (context_att);

        \node [var] (rt) [right of=context_att, xshift=0.5cm] {$r^{(t)}$};
        \draw [arrow] (context_att.east) -- (rt.west);
        



        \node [feat, minimum height=2.0cm] (dec_mlp) [right of=dec_fac, xshift=2.6cm, yshift=0.1cm] {$f_y$};
        \draw [arrow] (rt.east) -- (rt.east -| dec_mlp.west);
        \draw [arrow] (h_t_dec.east) to [out=0,in=180] (dec_fac.east -| dec_mlp.west);

        \node [realis_prior] (z_C) [right of=y_feat_C, xshift=0.6cm] {$z_C$};
        \draw [arrow_test] (z_C.south) |- node[pos=0.25,name=5]{} ([yshift=-0.7cm]dec_mlp.west);

        \node [var_pred] (y_pred_t) [right of=h_t_dec, xshift=2.7cm] {$y^{(t)}_T$};

        \draw [arrow] (y_pred_t.west -| dec_mlp.east) -- (y_pred_t.west);
        

    \end{scope}

    \begin{pgfonlayer}{background}
        \filldraw [line width=3mm,join=round,DecCellColor]
            ([yshift=0.1cm]context_att.north  -| dec_mlp.east) rectangle ([xshift=-0.1cm] dec_mlp.south  -| dec_fac.west);
    \end{pgfonlayer}


\end{tikzpicture}
        \caption{Decoder cell $f_\mathsf{dec}$}
        \label{fig:dec_arch}
    \end{subfigure}%
    
    \caption{\footnotesize Computation diagram of the generative process of CCHP. In (a), the encoder (left) represent the posterior $q_\phi$ conditioned on context. ``$\sim$'' means sample operation. The decoder (right) represents the generative model $p_\theta(\by_T\mid \bx_T,\bx_C,\by_C,z)$. Both encoder and decoder have a recurrent structure. The structure of each individual cell is shown in (b) and (c), shaded with matching color ($f_\mathsf{enc}$ in purple and $f_\mathsf{dec}$ in cyan). $f_\mathsf{hand}$ in (b) is a learnable function.}
    \label{fig:model_arc}
    \vspace{-16pt}
\end{figure*}

\subsection{Learning and inference of CCHP}\label{sec:learn_infer_CCHP}

To learn the distribution \eqref{eq:generative_process} from data, we approximate the posterior of $z$ using a variational distribution $q_\phi(z\mid \bx_T,\by_T,\bx_C,\by_C)$ and minimize its Kullback–Leibler (KL) divergence with the true posterior, given by
\begin{equation}\label{eq:KL}
    \minimizewrt{\phi} D_{\mathsf{KL}}(q_\phi(z\mid \bx_T,\by_T,\bx_C,\by_C) \mid\mid p(z\mid \bx_T,\by_T,\bx_C,\by_C))
\end{equation}
where $\phi$ parameterizes the varational posterior $q$. It can be shown that solving \eqref{eq:KL} is equivalent to maximizing the following evidence lower bound (ELBO):
\begin{align}
    & \log p(\by_T\mid \bx_T,\bx_C,\by_C) \geq \mathsf{ELBO}\label{eq:elbo_1}
\end{align}
where $\mathsf{ELBO}$ is given by
\begin{align}
    \mathsf{ELBO} &\defeq \Expect{z\sim q_\phi(z\mid \bx_T,\by_T,\bx_C,\by_C)}{\log p_\theta(\by_T\mid \bx_T,\bx_C,\by_C,z)} \nonumber \\
    & - D_\mathsf{KL}(q_\phi(z\mid \bx_T,\by_T,\bx_C,\by_C) \mid\mid p(z)). \label{eq:ELBO_def}
\end{align}
See the online Appendix A\footnote[1]{\codeurl} for a detailed derivation of \eqref{eq:elbo_1} and \eqref{eq:ELBO_def}.
Now, note that to optimize \eqref{eq:ELBO_def}, we need to know the prior $p(z)$, which is usually either intractable or assumed to be known (e.g., standard Gaussian).
In this work, we follow \cite{attitude_2017} to approximate it using the variational posterior $q_\phi(z\mid \bx_C,\by_C)$ conditioned on context data only.
Instead of using an uninformed prior as in the case of variational autoencoders \cite{kingma2014autoencoding}, we are extracting useful information about each user from a personalized database.
The KL term in \eqref{eq:ELBO_def} then essentially keeps the posterior distribution consistent within each user, conditioned on either context (past interaction) or target (new interaction).
This can be more clearly seen in the following ELBO form:
\begin{align}
    \mathsf{ELBO}=&\sum_{t=1}^{N_T} \Expect{z\sim q_{\phi\mid T}}{\log p_\theta(y_T^{(t)}\mid \by_T^{(1:t-1)}, x_T^{(t)},\bx_C,\by_C,z)} \nonumber \\
     - D_\mathsf{KL} & (q_\phi(z\mid \bx_T,\by_T,\bx_C,\by_C) \mid\mid q_\phi(z\mid \bx_C,\by_C)). \label{eq:elbo_2}
\end{align}
where $q_{\phi\mid T}$ abbreviates $q_\phi(z\mid \bx_T,\by_T,\bx_C,\by_C)$. Notice that in \eqref{eq:elbo_2}, we also incorporate the temporal structure of $\by_T$ by expanding $p_\theta(\by_T\mid \bx_T,\bx_C,\by_C,z)$ for each time step.

To practically solve \eqref{eq:elbo_2}, we assume that both the generative model $p$ and the approximate inference model $q$ are Gaussian distributions that can be parameterized by learnable functions, e.g., artificial neural networks.
Specifically, we assume a parameterized Gaussian posterior $q_\phi(z\mid\bx_*,\by_*)$ as $z\sim\mathcal{N}\left(\mu_\phi,\Sigma_\phi\right)$.
We assume $\mu_\phi, \Sigma_\phi = F_\mathsf{enc}(\bx_*,\by_*\mid \phi)$ where $F_\mathsf{enc}$ is a nonlinear \textit{encoder} function parameterized by $\phi$.
When $(\bx_*,\by_*)$ involves target data, we refer to the resulting distribution \textit{latent posterior}, or $q_{\phi\mid T}$ as in \eqref{eq:elbo_2}.
When $(\bx_*,\by_*)$ contains only context data, we have the \textit{approximate latent prior}, or $q_{\phi\mid C}\defeq q_\phi(z\mid \bx_C,\by_C)$.
Similarly, we assume the generative process $p_\theta(y_T^{(t)}\mid \by_T^{(1:t-1)}, x_T^{(t)},\bx_C,\by_C,z)$ as $y_T^{(t)}\sim\mathcal{N}\left(\mu_\theta^{(t)},\Sigma_\theta^{(t)}\right)$ where $\mu_\theta^{(t)}$ and $\Sigma_\theta^{(t)}$ are generated by a neural network \textit{decoder} $F_\mathsf{dec}(\by_T^{(1:t-1)}, x_T^{(t)},\bx_C,\by_C,z\mid \theta)$. Now, we can write the goal of cobot handling as the following:
\begin{equation}
    \minimizewrt{\phi,\theta} \sum_{t=1}^{N_T} \Expect{z\sim q_{\phi\mid T}}{\log p(y_T^{(t)}\mid \mu_\theta^{(t)},\Sigma_\theta^{(t)})} - D_\mathsf{KL} (q_{\phi\mid T} \mid\mid q_{\phi\mid C}). \label{eq:elbo_3}
\end{equation}

The above problem can be solved using gradient-based methods.
By solving \eqref{eq:elbo_3}, we can improve inference on the latent $z$ while learning the generative model $p_\theta(\by_T\mid \bx_T,\bx_C,\by_C,z)$ from offline data $D_\mathsf{train}$.
During online deployment, we compute cobot action by sampling from the learned generative model $p_\theta$, where the latent $z$ is sampled from the approximate latent prior $q_{\phi\mid C}$.

\vspace{-15pt}


\subsection{Neural Network Architecture of CCHP}\label{sec:arch}



The structure of CCHP is shown in \cref{fig:CCHP_arch}.
The encoder $q_\phi$ takes any dynamic gestures $\bx$ and corresponding handling operations $\by$ as input and generates a Gaussian posterior of latent $z$.
The decoder $p_\theta$ takes current gestures $\bx_T$ and sampled $z$ as input and generates a Gaussian likelihood for the desired handling operation $\by_T$, conditioned on some context information.
Note that \cref{fig:model_arc} corresponds to the testing time where we sample $z$ from approximate prior $q_\phi(z\mid \bx_C,\by_C)$ in the generative process.
For the posterior in \eqref{eq:elbo_3}, we feed both context and target to the encoder and get $q_\phi(z\mid \bx_T,\by_T,\bx_C,\by_C)$.
In the next section, we describe the encoder in terms of a general form of input $(\bx,\by)$ for simplicity.

\subsubsection{Encoder $q_\phi$}\label{sec:cchp_encoder}

A complete diagram of the encoder is shown in \cref{fig:CCHP_arch} (left).
Given an input cobot handling trajectory $(\bx,\by)$,
the encoder first computes a hidden state $\bh$ that encodes the input dynamic gesture command at each time step.
Here, $\bh\defeq (h^{(i)})_{i=1}^N$ where $h^{(t)}\in\RR^{H}$ and $H$ is the hidden state size.
To introduce temporal dependency in $\by$ (see Eq.~\eqref{eq:generative_process}),
we incorporate a recurrent structure (see \cref{fig:CCHP_arch}) as a concatenation of \textit{encoder cells} ($f_\mathsf{enc}$).
The encoder cell, as shown in \cref{fig:enc_arch}, interprets the current user command $x^{(t)}$ into a hidden state $h^{(t)}$ while considering $y^{(t-1)}$ and $h^{(t-1)}$.
With the feature $\bh$ extracted from input data, we summarize $\bh$ and $\by$ using an aggregation function $a$, and then generate the final Gaussian posterior $q_\phi(z\mid \bx,\by)$ with mean $\mu_\phi$ and covariance $\Sigma_\phi$.
This procedure follows \cite{anp} and can be summarized as:
\begin{align}
    \mu_\phi,\Sigma_\phi = f_\mathsf{\phi}(a((h^{(1)},y^{(1)}),\dots,(h^{(N)},y^{(N)}))), \label{eq:sC_fa_f_phi}
\end{align}
where $a$ can be any function that reduces the time dimension and is chosen as mean function in this work.
$f_\phi$ is a learnable function that produces the final statistics of latent $z$, assuming $\Sigma_\phi$ is a diagonal matrix.

\subsubsection{Decoder $p_\theta$}\label{sec:cchp_decoder}

Using the context $\bh_C$ and $\by_C$ and latent $z_C$, the decoder predicts the handling operations $\by_T$ for a target commands $\bx_T$.
The explicit dependency of $\Pi_\theta$ on user context (see section \ref{sec:cchp_intro}) is encoded in $(\bh_C,\by_C)$ pair, while the temporal dependence in $\by_T$ is captured by a recurrent structure similar to that in the encoder.
The basic unit of the decoder is a \textit{decoder cell} which processes one target command $x_T^{(t)}$ at a time to predict $y_T^{(t)}$, considering $\by_T^{(1:t-1)}$ and user context.

 
Shown in \cref{fig:dec_arch}, the decoder cell first invokes the encoder cell $f_\mathsf{enc}$ to produce
$h_T^{(t)}$.
It is then fed to a \textit{context attention} module to generate a representation $r^{(t)}$ that captures most relevant context for current time step.
Context attention, as in \cite{anp}, learns to attend over context hidden states $\bh_C$, known as keys, for each target hidden state $h^{(t)}_T$, known as a query.
This is done by selecting similar context-target states in dot product sense and computing weights over the context $\by_C$:
\begin{align}\label{eq:context_att}
    \lambda_{u}^{(t)} &= \mathrm{softmax}(\langle h_C^{(u)}, h^{(t)}_T\rangle),~\forall u\in[N_C].
\end{align}
Then, we have $r^{(t)} = \sum_{u=1}^{N_C} \lambda_{u}^{(t)} y^{(u)}_C$.
Intuitively, the attention module is a similarity measure for dynamic gestures and extract insights from user-specific data to support online prediction.
Finally, we predict ${y}_T^{(t)}$ as
$\mu_\theta^{(t)}, \Sigma_\theta^{(t)} = f_y(r^{(t)}, h_T^{(t)}, z_C)$.

Note that, CCHP is derived based on novel intuitions with sound theoretical formulation.
Nevertheless, the NN architecture needs to be correctly designed to represent the correct function class in order to effectively solve real-world applications.
Our design serves as a practical reference for other tasks when similar features are desired. See the online Appendix C\footnotemark[1] for more architecture details.

\subsubsection{Training CCHP with Teacher Forcing}

According to \eqref{eq:generative_process}, when predicting $y_T^{(t)}$, we should use ground truth $y_T^{(t-1)}$ from the previous step.
If we do so, the model will always assume that its previous output is correct.
At test time, however, the model will inevitably make mistakes in previous timesteps.
Such error propagation will build up and lead to a large covariate shift between the training and online distribution over $\by_T^{(1:t-1)}$.
This would ultimately cause the model to fail during testing.
As a remedy, we apply a practical modification to the training procedure called \textit{teacher forcing}, initially noted by \cite{bengio2015teacherforce} and popularized by \cite{chan2015listen}. This technique allows the model's own prediction $\hat{y}_T^{(t-1)}$ to be fed to next time step with some probability $1 - p_\mathsf{TF}$, shown as the additional green cell in \cref{fig:dec_arch}. The probability $p_\mathsf{TF}$ of using ground truth gradually decreases throughout training. With that, the model initially learns to generate reasonable handling operations in short term, and then work on long-term prediction to gain robustness against its own previous errors. In our model, we reduce $p_\mathsf{TF}$ at a fixed and linear rate.

\vspace{-12pt}




\section{Technical Evaluation}
\label{sec:eval}


We introduce the data collection (section \ref{sec:collect_user_policies}), present quantitative evaluations (section \ref{sec:quant_eval}) on different models (section \ref{sec:baseline_ablation_eval_metric}), and interpret our model in section \ref{sec:qual_result}.

\vspace{-15pt}

\subsection{Collection of Human-Human Handling Demonstrations}
\label{sec:collect_user_policies}

To collect training and testing data for CCHP model, we record human-human collaborations.
The user performs dynamic gestures $\bx$ to achieve the object motions as indicated by simulated animations, while the corresponding handling operations $\by$ are labeled by another human (see \cref{fig:rtcohand_info}).
The two humans can communicate to ensure that the user is satisfied with the labeled operations.
Moreover, we need each user to have a consistent style (\cref{sec:formulation_derive}) for our model to learn well.
In this work, we verify the consistency by conducting a t-test for each user on whether the pair-wise dynamic time warping (DTW) \cite{nd_dtw} distances between the $\bx$'s for the same $\by$'s are significantly smaller than those between the $\bx$'s for different $\by$'s.
We consider a user to be consistent if $p<.0001$, or otherwise reject the user data.
The approach to data quality assurance can be tuned to reflect different consistency requirements in future work.

\begin{figure}[t]
    \centering
    \includegraphics[width=0.9\linewidth]{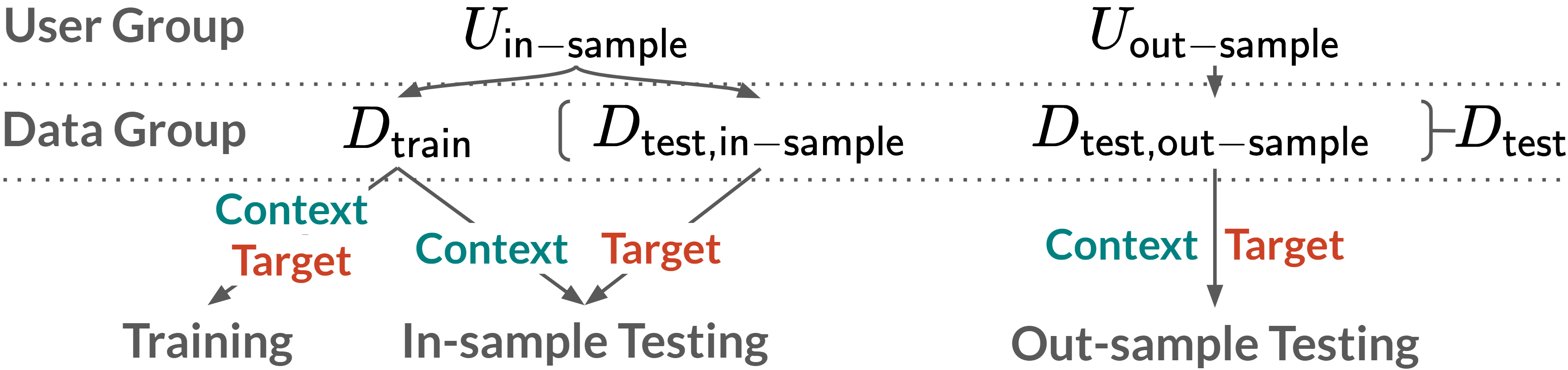}
    \caption{Relations between user, data, and train test settings.
    }
    \label{fig:user_data_relation}
    \vspace{-15pt}
\end{figure}

We collect human-human collaboration with $15$ users\footnote[2]{The data collection process only involves two humans performing hand motions and maneuvering a light-weight paper box. The harm or discomfort anticipated is no greater than those ordinarily encountered in daily life.} in total.
We assign $10$ users to in-sample user group ($U_\mathsf{in-sample}$) and divide their data into a training set $D_\mathsf{train}$ and testing set $D_\mathsf{test, in-sample}$ (see the online Appendix B\footnotemark[1] for details).
When testing with $D_\mathsf{test, in-sample}$, we sample $(\bx_T, \by_T)$ from $D_\mathsf{test, in-sample}$ and $(\bx_C, \by_C)$ from $D_\mathsf{train}$.
This replicates the cases where the user's data has appeared in training.
The rest $5$ users belong to out-sample user group ($U_\mathsf{out-sample}$) whose data make the test set $D_\mathsf{test, out-sample}$.
When testing with $D_\mathsf{test,out-sample}$, we draw both context and target samples from $D_\mathsf{test, out-sample}$.
This replicates the cases where a new user provides a short demonstration and work with our system.
See \cref{fig:user_data_relation} for an illustration.
It takes around one hour to collect training data from each user as a one-time effort.
When adapting to a new user during deployment (e.g., our hardware experiment to be covered in \cref{sec:user_study_exp_design}), it takes less than five minutes to collect the necessary context.

\vspace{-15pt}

\subsection{Baselines, Ablations, and Evaluation Metrics}\label{sec:baseline_ablation_eval_metric}
We compare our model's performance with three state-of-the-art baseline models:
(a) \textbf{motion cloning}~\cite{hand_to_end_effector} (MC) which mirrors the hand motion to the robot end-effector,
(b) \textbf{LSTM}~\cite{LSTM}, a two-layer recurrent neural network,
and (c) \textbf{RANP}~\cite{qin_recurrent_2019}, a standard ANP~\cite{anp} extended with recurrent feature extraction.
We include MC to test rule-based methods and LSTM to test learning-based methods.
Both MC and LSTM are context-free.
We implement RANP to test context-aware methods without temporal dependencies in the output.
The graphical representation of RANP is similar to \cref{fig:pgm} but without the connections between $y_T$'s.
We also analyze the impact of training procedures and test ablations in two aspects: (a) the probability $p_{\mathsf{M}}$ of the target motion appearing in context
and (b) the probability $p_\mathsf{TF}$ of using ground truth for previous steps in teacher forcing.
Our main CCHP model is trained with $p_\mathsf{TF}$ starting from $0.9$, staying fixed for the first 600 training steps, and then linearly decreasing.
$p_{\mathsf{M}}$ is constant $0.5$.
Our model training time is around five hours on a desktop with an RTX 2080Ti GPU and an Intel i9-9940X CPU.
More training details can be found in the online Appendix D\footnotemark[1].

We evaluate the model performance on three real-world use cases:
(1) \textbf{matching context}, using $D_\mathsf{test, in-sample}$ with target motion appearing in the context;
(2) \textbf{mismatching motion}, using $D_\mathsf{test, in-sample}$ with target motion not in the context;
and (3) \textbf{new user}, using $D_\mathsf{test, out-sample}$.
Finally, we examine whether our CCHP model is robust against \textbf{noisy human input} as motivated in section \ref{sec:cchp_intro}.
To do that, we perturb the target dynamic gestures $\bx_T$ with zero-mean Gaussian noises with $\sigma_T = 0.005$ m and $\sigma_R = 0.025$ rad for translation and rotation respectively on top of the matching context setting.

\vspace{-15pt}

\begin{figure}[t]

\centering

\resizebox{\linewidth}{!}{
\begin{tabular}{lcccc}
    \toprule
    Models & Matching & 
                 Mis-motion &
                 New user & 
                 Noisy\\
    \midrule
    Motion cloning~\cite{hand_to_end_effector} & (4.5, 8.814) &  NA & (4.4, 8.499) & (5.9, 12.626)\\
    
    LSTM & (3.9, 10.851) &  NA & (4.5, 12.328) & (5.1, 12.411)\\
    
    RANP~\cite{qin_recurrent_2019}  & (4.1, 6.190) & (4.3, 6.851) & (4.6, 7.307) & (6.1, 13.603)\\
    
    CCHP & (\textbf{3.7}, \textbf{5.776}) & (4.0, 6.322) & (\textbf{4.1}, \textbf{6.830}) &  (4.4, 8.585)\\
    
    CCHP $p_{\mathsf{M}}=0.1$  & (3.8, 5.859) & (4.0, \textbf{6.310}) & (4.2, 7.038) & (4.4, \textbf{8.247})\\
    
    CCHP $p_{\mathsf{M}}=1.0$ & (4.2, 6.236) & (5.5, 7.312) & (5.1, 7.586) & (4.7, 8.313)  \\
    
    
    CCHP $p_\mathsf{TF}=0.1$ & (3.7, 6.001) & (\textbf{3.9}, 6.597) & (4.1, 7.276) & (\textbf{4.3}, 10.379)\\

    CCHP $p_\mathsf{TF}=0.5$ & (3.8, 6.241) & (4.0, 6.756) & (4.1, 7.474) & (4.6, 9.844)\\
    
    CCHP $p_\mathsf{TF}=0.9$ & (7.7, 22.078) & (7.8, 22.338) & (10.2, 20.161) & (10.6, 20.473) \\
    \bottomrule
\end{tabular}
}
\captionof{table}{\footnotesize Test losses of different models (rows) under different test settings (columns).
``CCHP'' is our main model.
All results are shown as RMSE loss in translational velocity (cm/s) and mean rotational velocity error (degree/s) across all timesteps, the lower the better.
The best values appear in bold.
}
\label{table:loss}
\vspace{-16pt}
\end{figure}

\subsection{Empirical Quantitative Evaluation}\label{sec:quant_eval}

Table \ref{table:loss} compares model losses in terms of velocity errors on different test settings.
Comparing columns, the loss is consistently lower when the target motions appear in context, meaning that providing relevant context data does improve performance.
Performance on new user data is worse as expected since these users' hand motion policies were never observed during training.
The CCHP model achieves the best accuracy for new user data, showing the advantage of adaptation.
With input noises, we notice worse performance across all models, but our CCHP variants show the most robustness.
Cosidering ablations, all $p_\mathsf{TF}$ values except $0.9$ lead to similar scores to our main model.
This confirms the importance of exposing the model to its own errors during training.
The model is not sensitive to $p_\mathsf{M}$ values, but performs better when motion mismatch is present during training.
Comparing CCHP and baselines, both motion cloning and LSTM fail due to lack of adaptation.
RANP performs similarly to CCHP in all settings except noisy input due to the lack of temporal structure.

\begin{figure}[t]
    \centering
    \resizebox{1.0\linewidth}{!}{\input{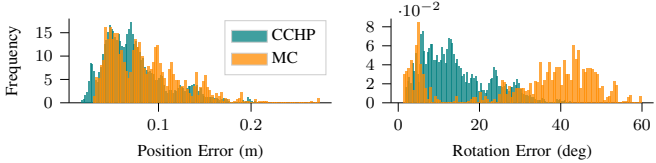}}
    \caption{Distribution of top-10\% pose errors under the “Matching” setting.}
    \label{fig:cumulative_error}
    \vspace{-10pt}
\end{figure}

The above velocity-based analysis shows the advantage of our method regarding instantaneous cobot behaviors.
To fully understand the significance of such improvement in practice, it is also essential to investigate the absolute cobot poses.
Unfortunately, since velocity errors can either stack or cancel with each other depending on the input, a pose-based metric cannot be readily obtained from the output velocities.
Moreover, pose errors can be dominated by task duration, and hence are not as instructive as velocity errors for general tasks.
Hence, we only provide an exemplary analysis specific to our experiment setting.
From \cref{fig:cumulative_error}, we find that the top $10\%$ cumulative rotation errors within $5$-second interactions concentrate at $10$ degrees with CCHP and at $40$ degrees with motion cloning.
This shows that improvements in velocity errors can accumulate and lead to drastically better robot poses over time.
The significance of such improvement is also evident in real-world applications (to be covered in \cref{sec:user_study}).

\vspace{-14pt}

\begin{figure}[t]
    \centering
    \includegraphics[width=0.85\linewidth]{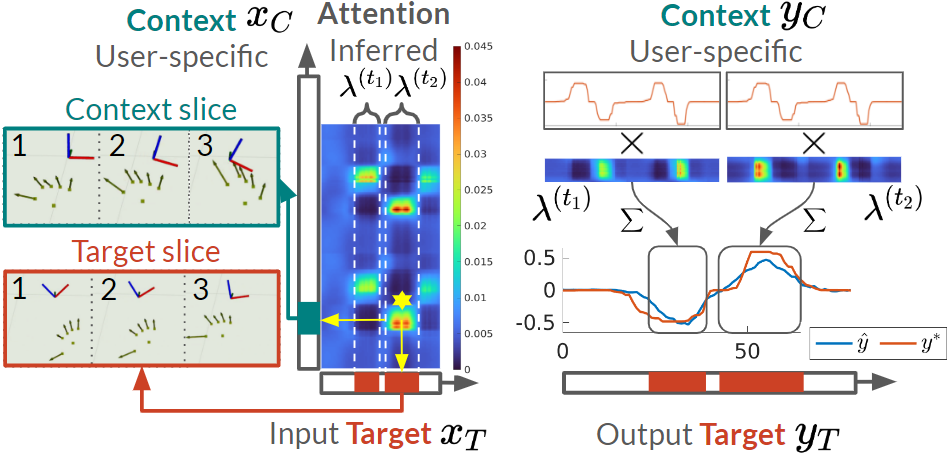}
    \caption{\footnotesize Computation diagram of CCHP with data visualization.
    (a) visualizes the attention weights (color map) of target dynamic gestures $\bx_T$ over context $\bx_C$.
    (b) shows the computation of target object motion $\by_T$ based on context $\by_C$.
    In (a), the axis represents the object pose with X, Y, Z axes in red, green, and blue respectively.
    In (b), the ground truth target operation is shown in red.
    All plots only show the rotation velocities in Y-axis.
    The marked regions on the horizontal axes in (a) and (b) refer to the same time ranges in target dynamic gestures (input) and handling operations (output) respectively.
    }
    \label{fig:qualitative_cases}
    \vspace{-15pt}
\end{figure}

\subsection{Empirical Qualitative Results}\label{sec:qual_result}



We now qualitatively interpret the computation of CCHP model.
We choose a case where both context and target operations contain Y-axis rotations.
Recall that the model computes context attention $\blambda\defeq \{\lambda^{(t)}_u\}_{t\in[N_T],u\in[N_C]}$ for each target time step $t$ over each context time step $u$ (see \cref{eq:context_att}).
The resulting weights $\blambda$ are shown in \cref{fig:qualitative_cases} (left).
One region of high attention is marked by a yellow star.
Visualizing the corresponding dynamic gestures, we see that they are indeed similar: in both cases, the hand is rotating clockwise, leading to Y-axis rotations in desired handling operations.

With $\blambda$, the model proceeds to generate target handling operation $y_T^{(t)}$.
We focus on two target time ranges marked in red in \cref{fig:qualitative_cases} and their context attentions $\lambda^{(t_1)}$ and $\lambda^{(t_2)}$.
As shown in \cref{fig:qualitative_cases} (right), the model calculates a weighted sum of $\by_C$ using $\lambda^{(t_1)}$ and $\lambda^{(t_2)}$ and generate $\by_T$.
The target output is highly related to the context at locations with high attention.
Here, Y-axis rotation is the only primary motion.
Our model is also able to compose full rigid body motions by attending to both translation and rotation context motions simultaneously.

\vspace{-10pt}

\section{User Study}
\label{sec:user_study}

We conduct a user study to investigate the benefit of using context in gesture-based cobot handling.
Each user first demonstrates a few dynamic gesture commands, and then assembles a piece of furniture with the help of a cobot handler.

\vspace{-10pt}

\subsection{Experiment Design}\label{sec:user_study_exp_design}

\textbf{Independent Variables.}
We manipulate the cobot handling policy with two levels: \textit{with context} and \textit{no context}.
The robot either interprets human gestures using CCHP model with user context or motion cloning (MC) which ignores user context and runs in fixed rule-based manners.

\textbf{Dependent Measures.}
In order to make the task objectives consistent among participants, we design and convey the desired task procedures and robot behaviors.
The participants are instructed to achieve those behaviors with minimum effort.
For each trial, we measure the human effort related to cobot handling in terms of total hand translation, finger rotation, and interaction time.
We also subjectively measure the smoothness of human-robot collaboration using 2 multi-item scales shown in \cref{tab:user_study}: \textit{do participants think that the robot is understanding their intentions} and \textit{how easy is it to collaborate with the robot}.

\begin{figure}[t]
    \centering
    \includegraphics[width=0.9\linewidth]{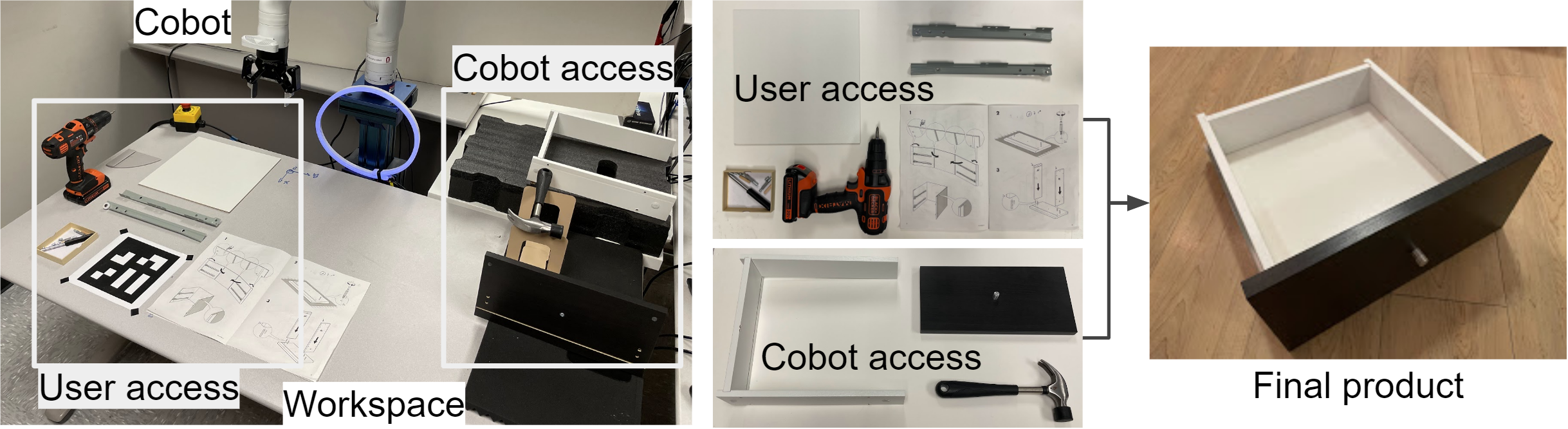}
    \caption{Furniture assembly task. Users have direct access to part of the workpieces and tools, and rely on the robot to handle the rest.}
    \label{fig:user_study}
    \vspace{-19pt}
\end{figure}

\textbf{Hypothesis H1.}
\textit{The use of context reduces human effort.}

\textbf{Hypothesis H2.}
\textit{Participants believe the robot understands their gestures and collaborates better with user context.}

\begin{table*}
    \footnotesize
    \centering
    \resizebox{0.9\linewidth}{!}{
\begin{tabular}{llccccc}
    \toprule
     & Questions & Cronbach's $\alpha$ & MC LSM & CCHP LSM & W & p-value\\
    \midrule
    
    \textbf{understanding} & \begin{tabular}{@{}l@{}} The robot actions exactly match my expectation. \\ The robot understands my command style. \\ The robot is able to ignore my unintentional/spasmodic motions. \end{tabular} & 0.84 & 3.83 & 5.77 & 2 & \textbf{.006} \\
    
    \midrule
    
    \textbf{collaboration} & \begin{tabular}{@{}l@{}} It is easy to collaborate with the robot. \\ I do not have to alter my gestures to achieve desired results. \\ The robot collaborated with me to complete the task. \end{tabular} & 0.79 & 4.53 & 6.00 & 3 & \textbf{.010} \\

    \bottomrule
\end{tabular}
}
    \caption{Results of subjective measures from a 7-point Likert-scale survey.}
    \label{tab:user_study}
    \vspace{-17pt}
\end{table*}

\textbf{Main Task.}
We design a realistic main user task: \textit{collaborative furniture assembly} (see \cref{fig:user_study}).
The user needs to assemble a furniture from components and fasteners following a pre-defined procedure.
In this process, a cobot is available to aid the user\footnote[3]{The cobot movement speed is highly constrained. The harm or discomfort anticipated is no greater than those ordinarily encountered in daily life.}.
We assign two cobot tasks: \textbf{(a)} tool management where the cobot fetches tools from the toolbox, and \textbf{(b)} object handling where the cobot holds a part of the furniture for the user to work on.
Both tasks follow a sequence of stages: \textbf{(i)} user pointing to a desired object, \textbf{(ii)} cobot passing the object to user, \textbf{(iii)} user optionally adjusting the cobot pose for comfortable working conditions, \textbf{(iv)} user optionally working on the held object (see \cref{fig:task_diagram}).
In this task, a cobot can greatly ease human effort and improve the efficiency as an additional hand.
As introduced at the beginning of this paper, cobot handling is intended only for ``last-millimeter'' personalized adjustments.
Hence, the user would benefit from not wearing specialized devices which could tamper the main task.
We implement the pipeline using OpenPose\cite{openpose} library for perception and a Kinova Gen3 robot for actuation.
Transition of the task stages is indicated by an LED ring.
The ring is progressively filled with stage-dependent colors when user inputs are detected until the transition completes.
The ring is fully on during \textbf{(ii)}, and fades away as the cobot policy times out in \textbf{(iii)}.

\textbf{Participants.}
We invite $10$ participants from the Carnegie Mellon community from different majors and randomize the order of the cobot policy conditions.

\textbf{Procedure.}
We first collect five-minute demonstrations from each participant.
Then, as familiarization, we show the assembly procedure and allow participants to practice cobot handling with a dummy cobot policy.
The participants then assemble the furniture once with each cobot policy condition and fill out the survey.

\vspace{-15pt}

\begin{figure}[t]
    \centering
    \begin{subfigure}[b]{0.6\linewidth}
    \centering
        \resizebox{1.0\linewidth}{!}{
\begin{tikzpicture}

    \definecolor{darkgray176}{RGB}{176,176,176}
    \definecolor{darkorange}{RGB}{255,140,0}
    \definecolor{lightgray204}{RGB}{204,204,204}
    \definecolor{teal}{RGB}{0,128,128}
    
    \begin{groupplot}[group style={group size=3 by 1}]
    \nextgroupplot[
    legend cell align={left},
    legend style={
      fill opacity=0.8,
      draw opacity=1,
      text opacity=1,
      at={(0.03,0.97)},
      anchor=north west,
      draw=lightgray204
    },
    tick align=outside,
    tick pos=left,
    xlabel={Translation (m)},
    xmin=0.56, xmax=1.44,
    y grid style={darkgray176},
    ymin=0, ymax=2.5,
    ytick style={color=black},
    width=2.9cm, height=4.05cm, xmajorticks=false, font=\footnotesize, axis x line*=bottom, axis y line*=left
    ]
    \draw[draw=black,fill=teal] (axis cs:0.6,0) rectangle (axis cs:1,0.652070769345524);
    \addlegendimage{ybar,ybar legend,area legend,draw=black,fill=teal}
    
    \draw[draw=black,fill=darkorange] (axis cs:1,0) rectangle (axis cs:1.4,1.80808988345172);
    \addlegendimage{ybar,ybar legend,area legend,draw=black,fill=darkorange}
    
    \path [draw=black, semithick]
    (axis cs:0.8,0.274132624939803)
    --(axis cs:0.8,1.03000891375125);
    
    \path [draw=black, semithick]
    (axis cs:1.2,1.23545801636319)
    --(axis cs:1.2,2.38072175054026);
    
    \addplot [semithick, black]
    table {%
    0.8 0.652070769345524
    };
    \addplot [semithick, black]
    table {%
    1.2 1.80808988345172
    };
    
    \nextgroupplot[
    legend cell align={left},
    legend style={
      fill opacity=0.8,
      draw opacity=1,
      text opacity=1,
      at={(0.03,0.97)},
      anchor=north west,
      draw=lightgray204
    },
    tick align=outside,
    tick pos=left,
    xlabel={Rotation (deg)},
    xmin=0.56, xmax=1.44,
    y grid style={darkgray176},
    ymin=0, ymax=999,
    ytick style={color=black},
    width=2.9cm, height=4.05cm, xmajorticks=false, font=\footnotesize, axis x line*=bottom, axis y line*=left
    ]
    \draw[draw=black,fill=teal] (axis cs:0.6,0) rectangle (axis cs:1,378.238177746677);
    \addlegendimage{ybar,ybar legend, area legend, draw=black,fill=teal}
    
    \draw[draw=black,fill=darkorange] (axis cs:1,0) rectangle (axis cs:1.4,810.449089359114);
    \addlegendimage{ybar,ybar legend, area legend, draw=black,fill=darkorange}
    
    \path [draw=black, semithick]
    (axis cs:0.8,315.46639312517)
    --(axis cs:0.8,441.009962368183);
    
    \path [draw=black, semithick]
    (axis cs:1.2,686.973332332429)
    --(axis cs:1.2,933.924846385799);
    
    \addplot [semithick, black]
    table {%
    0.8 378.238177746677
    };
    \addplot [semithick, black]
    table {%
    1.2 810.449089359114
    };
    
    \nextgroupplot[
    legend cell align={left},
    legend style={
      fill opacity=0.8,
      draw opacity=1,
      text opacity=1,
      at={(0.03,0.97)},
      anchor=north west,
      draw=lightgray204
    },
    tick align=outside,
    tick pos=left,
    xlabel={Time (s)},
    xmin=0.56, xmax=1.44,
    y grid style={darkgray176},
    ymin=0, ymax=129,
    ytick style={color=black},
    width=2.9cm, height=4.05cm, xmajorticks=false, font=\footnotesize, axis x line*=bottom, axis y line*=left
    ]
    
    
    \draw[draw=black,fill=teal] (axis cs:0.6,0) rectangle (axis cs:1,32.025);
    \addlegendimage{ybar,ybar legend, area legend, draw=black,fill=teal}
    \addlegendentry{CCHP}
    
    \draw[draw=black,fill=darkorange] (axis cs:1,0) rectangle (axis cs:1.4,63.9);
    \addlegendimage{ybar,ybar legend, area legend, draw=black,fill=darkorange}
    \addlegendentry{MC}
    
    \path [draw=black, semithick]
    (axis cs:0.8,30.8928008125776)
    --(axis cs:0.8,33.1571991874224);
    
    \path [draw=black, semithick]
    (axis cs:1.2,56.0282149419588)
    --(axis cs:1.2,71.7717850580412);
    
    \addplot [semithick, black]
    table {%
    0.8 32.025
    };
    \addplot [semithick, black]
    table {%
    1.2 63.9
    };
    \end{groupplot}
    
    \end{tikzpicture}
    }
        \caption{Average total human effort.}
        \label{fig:effort}
    \end{subfigure}%
    \hfill
    \begin{subfigure}[b]{0.39\linewidth}
    \centering
        \includegraphics[width=\linewidth]{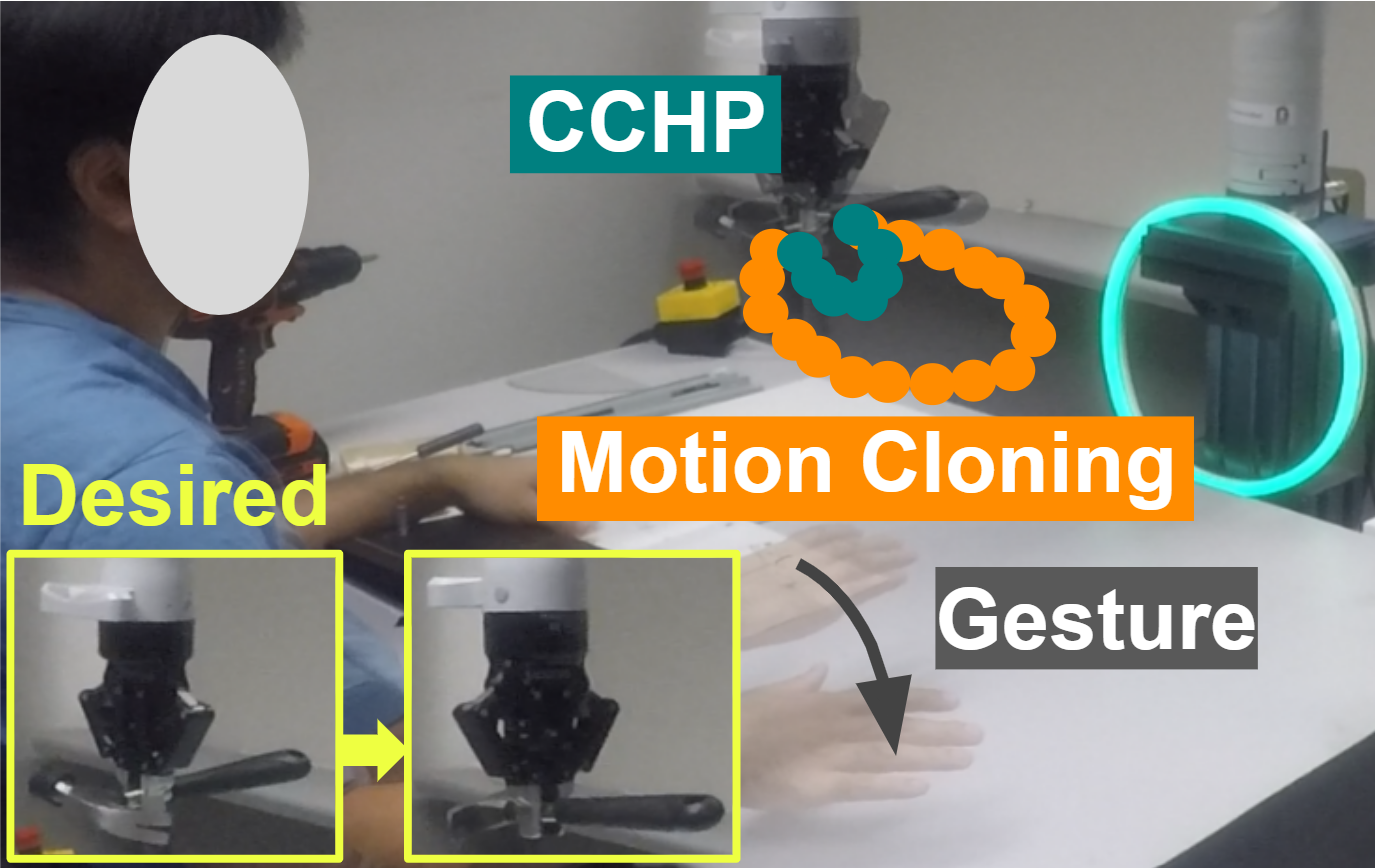}
        \caption{Sample participant trajectories.}
        \label{fig:traj}
    \end{subfigure}%
    \caption{(a) CCHP significantly reduces human effort in translation ($p=.003$), rotation ($p=.006$) and time ($p=.004$) compared to MC. (b) A participant desires an in-place rotation, which is captured by CCHP. MC misinterprets the gestures, leading to extra human effort to correct the cobot.}
    \label{fig:traj_effort}
    \vspace{-16pt}
\end{figure}

\subsection{Results}

For \textbf{objective measures}, we conduct a paired t-test and report results in \cref{fig:traj_effort}.
The use of context reduces human effort in translation $(t(9) = -8.59,~p = .003)$, rotation $(t(9) = -6.85,~p = .006)$ and time $(t(9) = -7.84,~p = .004)$ which supports H1.
We find very strong evidence that context is useful for understanding dynamic gestures.
For \textbf{subjective measures}, we conduct a Wilcoxon signed-rank test and report results in \cref{tab:user_study}.
Both scales are tested to be reliable.
The results suggest that with context (CCHP), the robot understands user input better ($W=2,~p=.006$) and leads to better collaboration ($W=3,~p=.010$).
H2 is also supported.
Hence, we conclude that context-aware cobots are significantly better helpers than those running in fixed rule-based manners.

\vspace{-5pt}
\section{Conclusion}
\label{sec:conclusion}

In this paper, we showed the benefit of using demonstrations as context for online understanding of dynamic gestures.
A probabilistic view of cobot handling was introduced to achieve robustness against noisy human inputs, which was validated on offline data.
Our user studies verified that with context, participants saved efforts and collaborated with the robot better.
Our work is a step toward fast adaptation in tasks with high-dimensional time sequence input.
We point out that our method might fail when the user command style during interaction shifts significantly from the context.
One potential mitigation is to accept online user corrections and adapt user context accordingly.
Another promising future direction is to incorporate object semantics into context for explicit adaptation to various tasks besides user styles.

\vspace{-8pt}


\bibliographystyle{IEEEtran}
\bibliography{ref}

\newpage

\appendix

\subsection{Minimization of KL divergence}\label{append:KL_ELBO}

We now show that minimizing the KL divergence \eqref{eq:KL} is equivalent to maximizing the ELBO as in \eqref{eq:elbo_1} and \eqref{eq:ELBO_def}. Recall the learning objective of CCHP:
\begin{equation}\label{eq:KL_append}
    \minimizewrt{\phi} D_{\mathsf{KL}}(q_\phi(z\mid \bx_T,\by_T,\bx_C,\by_C) \mid\mid p(z\mid \bx_T,\by_T,\bx_C,\by_C))
\end{equation}
First, we have
\begin{align}
    p(z\mid \bx_T,\by_T,\bx_C,\by_C) = \frac{p(\by_T\mid \bx_T, \bx_C, \by_C, z)p(z)}{p(\by_T\mid \bx_T, \bx_C,\by_C)}.
\end{align}
Plugging in the objective of \eqref{eq:KL_append} and expand, we have
\begin{align}
    D_\mathsf{KL} & = \EE_{z\sim q_\phi(z\mid \bx_T,\by_T,\bx_C,\by_C)}[\log q_\phi(z\mid \bx_T,\by_T,\bx_C,\by_C) \nonumber\\
    & - \log p(z) - \log p(\by_T\mid \bx_T,\bx_C,\by_C,z) \nonumber\\
    & + \log p(\by_T\mid\bx_T,\bx_C,\by_C)] \geq 0 \label{eq:KL_expand_append}
\end{align}
Rearranging, we have
\begin{align}
    & \log p(\by_T\mid\bx_C,\by_C,\bx_T) \nonumber\\
    \geq & \EE_{z\sim q_\phi(z\mid \bx_T,\by_T,\bx_C,\by_C)}[\log p(\by_T\mid \bx_T,\bx_C,\by_C,z) \nonumber\\
    &- \log q_\phi(z\mid \bx_T,\by_T,\bx_C,\by_C) + \log p(z)] \nonumber\\
    = & \EE_{z\sim q_\phi(z\mid \bx_T,\by_T,\bx_C,\by_C)}[\log p(\by_T\mid \bx_T,\bx_C,\by_C,z)] \nonumber\\
    & - D_\mathsf{KL}(q_\phi(z\mid \bx_T,\by_T,\bx_C,\by_C) \mid\mid p(z)) \triangleq \mathsf{ELBO} \label{eq:ELBO_append}
\end{align}
Plug \eqref{eq:KL_expand_append} and \eqref{eq:ELBO_append} into \eqref{eq:KL_append}, we have
\begin{align}
    & \minimizewrt{\phi} D_{\mathsf{KL}}(q_\phi(z\mid \bx_T,\by_T,\bx_C,\by_C) \mid\mid p(z\mid \bx_T,\by_T,\bx_C,\by_C)) \nonumber \\
    \equiv & \minimizewrt{\phi} -\mathsf{ELBO} + \log p(\by_T\mid\bx_T,\bx_C,\by_C) \nonumber \\
    \equiv & \maximizewrt{\phi} \mathsf{ELBO}
\end{align}
since $\log p(\by_T\mid\bx_T,\bx_C,\by_C)$ is intractable and does not depend on $\phi$. Parameterizing the likelihood $p(\by_T\mid \bx_T,\bx_C,\by_C,z)$ in \eqref{eq:ELBO_append}, we arrive at \eqref{eq:elbo_1} and \eqref{eq:ELBO_def}. $\blacksquare$

    







\subsection{Train Test Data Collection}\label{append:train_test_data}
We collect user demonstrations in short clips, each lasting for around $5$ seconds.
We collect in total $72$ clips from each user.
$24$ of the clips contain translation or rotation only (\textit{type 1}) with the remaining $48$ clips contain both translation and rotation (\textit{type 2}).
Clips of each type above are split evenly among $D_\mathsf{train}$ and $D_\mathsf{test, in-sample}$ for $U_\mathsf{in-sample}$.
Among the $24$ type 2 clips assigned to $D_\mathsf{test, in-sample}$, half contains motions that also appear in $D_\mathsf{train}$.


\subsection{Model Architecture}\label{append:model_arch}



The configurations of different MLP modules introduced in section \ref{sec:arch} are given in table \ref{table:model_arch_config}. Each configuration starts with input size and ending with output size. A ReLU activation is applied after input layer and every hidden layer. The hidden state size $H$ in section \ref{sec:cchp_encoder} is $128$.

\begin{table}[ht]\centering
    \begin{tabular}{lll}
        \toprule
        Module & Description & Configuration  \\
        \midrule
        $f_\mathsf{hand}$ &
        \begin{tabular}{@{}l@{}} Hand feature (\cref{sec:cchp_encoder}) \end{tabular}
         & [320, 128, 64, 32]\\
        \midrule
        $f_{\phi}$ & Latent posterior predictor in \cref{eq:sC_fa_f_phi} &
        \begin{tabular}{@{}l@{}} [128, 128, 128, 128, 128] \\ and two heads [128, 32] \end{tabular} \\
        \midrule
        $f_y$ & Operation prediction in \cref{sec:cchp_decoder} &   \begin{tabular}{@{}l@{}}[156, 128, 64, 2], one for \\ each motion dimension  \end{tabular} \\
        \bottomrule
    \end{tabular}
    \caption{MLP configurations for CCHP neural network implementation.}
    \label{table:model_arch_config}
\end{table}

\subsection{Model Training}\label{append:model_training}
We train each model using the Adam optimizer with learning rate $lr = 5e-4$. During training, command finger velocities $\bx$ are perturbed with Gaussian noises $\epsilon \sim \cN(0, 1e-6)$. We train our models for a total of $738$ epochs with batch size $32$, which took around $5$ hours using an Nvidia GeForce RTX 2080Ti GPU with an Intel i9-9940X CPU.








\end{document}